\documentclass{article}

\usepackage[accepted]{icml2025}

\usepackage{microtype}
\usepackage{graphicx}
\usepackage{booktabs}
\usepackage{hyperref}
\usepackage{amsmath}
\usepackage{amssymb}
\usepackage{mathtools}
\usepackage{tikz-cd}
\usepackage{enumitem}

\usepackage{booktabs}
\usepackage{multirow}
\usepackage{array}
\usepackage{siunitx}
\usepackage{comment}
\usepackage{makecell}
\usepackage{graphicx}
\usepackage{subcaption}
\usepackage[compatibility=false]{caption}
\usepackage{longtable}
\usepackage{booktabs}  
\usepackage{placeins}

\usepackage{xcolor}

\makeatletter
\renewcommand{\ICML@appearing}{%
  \textit{Mechanistic Interpretability Workshop at the
  $\mathit{43}^{rd}$ International Conference on Machine Learning},
  Seoul, South Korea, 2026.
  Copyright 2026 by the author(s).%
}
\makeatother

\icmltitlerunning{LAWFUL: Law-Aligned Witness for Faithful Use of Latents}

\begin{document}

\twocolumn[
\icmltitle{ LAWFUL: Law-Aligned Witness for Faithful Use of Latents}

\begin{icmlauthorlist}
\icmlauthor{Kevin Chen}{inst1}
\icmlauthor{Kenneth W. Parker}{inst2}
\icmlauthor{Anish Arora}{inst1}
\end{icmlauthorlist}

\icmlaffiliation{inst1}{Computer Science \& Engineering, The Ohio State University, Columbus, Ohio, USA}
\icmlaffiliation{inst2}{Independent Researcher, Waterford, Virginia, USA}

\icmlcorrespondingauthor{Kevin Chen}{chen.11020@osu.edu}
\icmlkeywords{mechanistic interpretability, circuit discovery, physics-based interpretability, continuous physical variables}

\vskip 0.3in
]

\printAffiliationsAndNotice{}

\begin{abstract}

When a neural network predicts a physical system accurately, has it learned the governing law as formal, structured knowledge, and if so, does the network's internal computation actually use that representation throughout the law's domain of validity? We identify four interpretability gaps that limit answering these questions for {\em physics laws over continuous variables}: the absence of a coverage-aware causal-consistency measure over continuous counterfactuals; of a domain-of-validity test for the identified circuit; of a verification of the law's invariants and forbidden behaviors; and of a quantification of how a derived physical quantity flows through the circuit. We develop a foundational framework, LAWFUL, that closes the first two and lays groundwork for the remaining two, and illustrate it on the Mocap2Radar transformer, validating whether it learns and internally uses the Doppler frequency law $f(t) = \frac{2 v(t)}{\lambda}$
from motion-capture and radar data in which neither $f(t)$ nor $v(t)$ appears. Our source code is available at: \url{https://github.com/aciculachen/LAWFUL}



\end{abstract}

\label{sec:picc}

\section{Introduction}

{\em When a neural network predicts a physical system accurately, has it actually learned the governing physical law as formal, structured knowledge over the continuous variables that law involves? If so, is that representation causally used by the network's internal computation, or is the representation merely something an external probe can decode \citep{ belinkov2021probing, hewitt2019designing}? And does the representation hold throughout the domain of validity of the law, or only within the narrow distribution from which the training or validation data were drawn?} These are not three rephrasings of ``is the model accurate"; they are three distinct claims about what the network has internalized, and answering them requires mechanistic interpretability rather than behavioral evaluation, and a formal framework that can do so would furnish the principled basis on which the model can be trusted to behave lawfully wherever the law does.


Despite recent progress in mechanistically interpreting how neural models learn physical phenomena, the state of the art for laws expressed over continuous physical variables leaves several interpretability gaps that no existing framework closes. First, although causal abstraction with distributed alignment search and interchange-intervention accuracy \cite{geiger2025causal, geiger2024das, wu2023interpretability} supplies a formal notion of faithfulness, current instantiations quantify causal consistency only on narrowly sampled, discrete counterfactual pairs \citep{sutter2025nonlinear, makelov2024subspace, meloux2025everything}; they provide no measure designed to certify consistency over counterfactual families that cover the {\em continuous} space of the law's variables. Second, no published mechanistic procedure establishes whether an identified physics circuit operates consistently across the full range over which the law actually holds \citep{friedman2024illusions, nanda2023progress, vafa2025foundation, liu2026kepler}—that is, whether it generalizes over the law's {\em domain of validity} rather than over the empirical training or validation distribution. Circuits with provable guarantees over continuous input domains have been considered for robustness over input perturbations \citep{hadad2026formal}, but not for consistency with a known physics law over a derived-feature space.

\begin{figure}[t]
  \centering
  \includegraphics[width=\linewidth]{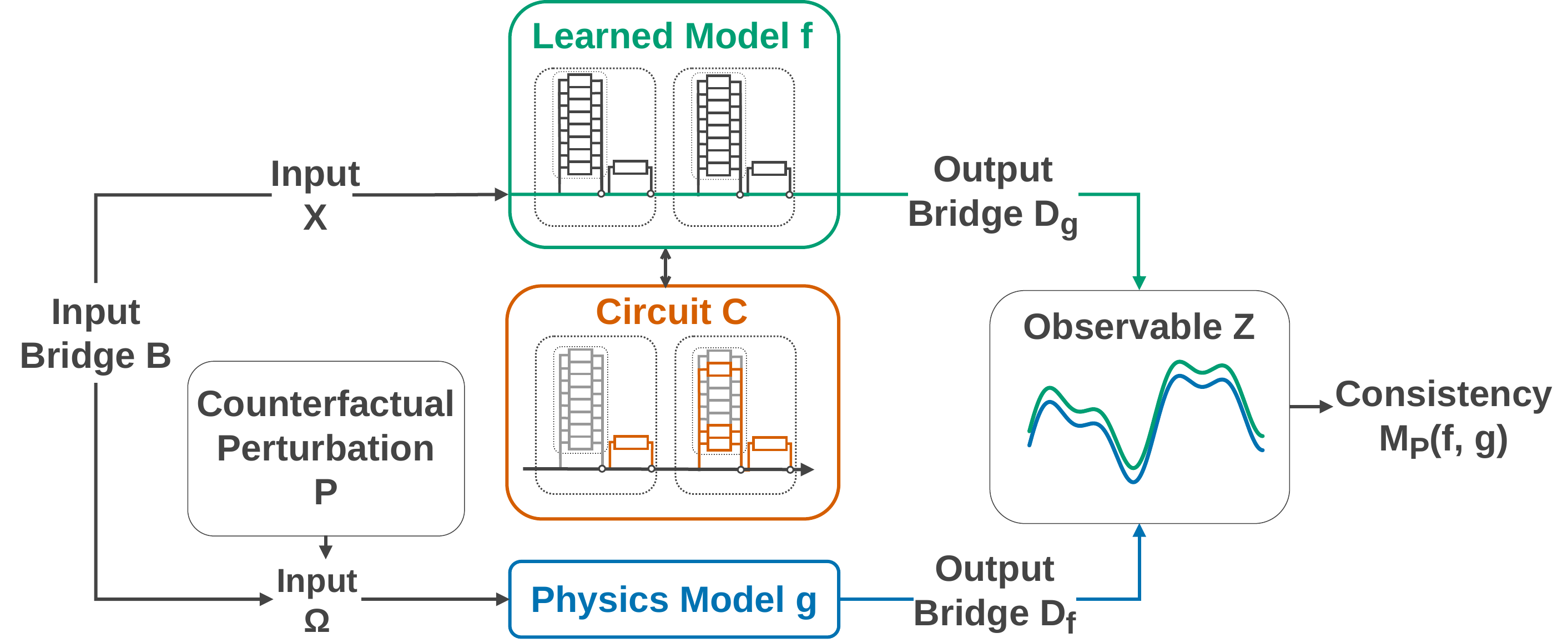}
    \caption{In LAWFUL, an input bridge maps the learned model input to the physics model input, which allows considering counterfactual perturbations over physically meaningful continuous domains. The output bridges project both models' outputs into a shared observable space, from which the physical consistency score is computed. This score serves as the metric for circuit identification and for checking invariances in the physics model.}
    \label{fig:framework}
\end{figure}

We develop a foundational treatment that closes both gaps within a single coherent formal framework, LAWFUL, supplying a coverage-aware causal-consistency measure defined over continuous counterfactual families together with a domain-of-validity test for the circuit; see Figure~\ref{fig:framework}. We illustrate the framework on a problem in radar physics: validating whether the sequence-to-sequence transformer MoCap-to-Radar~\citep{chen2026physics} explicitly learns, and internally uses, the Doppler frequency law, $f(t) = \frac{2 v(t)}{\lambda}$, where $f(t)$ and $v(t)$ are the Doppler frequency and radial velocity at time $t$, and $\lambda$ the radar wavelength.  The model maps three-dimensional motion-capture (MoCap) trajectories to radar spectrograms via a spatial transformer composed sequentially with a temporal transformer, trained on MoCap recordings of a human subject wearing 53 reflective markers paired with measurements from a Bumblebee~\cite{samraksh_bumblebee_product} homodyne radar. Crucially, neither $f(t)$ nor $v(t)$ appears in the training data, and the trained model generalizes to out-of-distribution targets — for example, random walks drawn from a class disjoint from the fixed-direction walks used during training.

The same framework lays the groundwork for redressing two further interpretability gaps, which we illustrate only partially through the MoCap-to-Radar model. The first concerns the {\em negative} content of a physical law, i.e., its associated invariants, conservation principles, and forbidden behaviors \citep{marks2025sparse, belrose2023leace, actionable2026symmetries, crabbe2023robustness}, for which no existing procedure verifies mechanistic enforcement inside the circuit, since concept-erasure and causal-scrubbing tools target spurious correlations rather than physically mandated invariances. The second concerns the {\em flow of the physical variables} themselves through the circuit: while path-patching, attribution-patching, and attribution-graph methods \citep{wang2023ioi, conmy2023acdc, syed2024attribution, heimersheim2024activation} identify which edges matter, they do not quantify how the continuous signal of a derived physical quantity is written, transformed, transmitted, and read by the heads and MLPs that constitute the law-implementing pathway. Combined with the causal-consistency measure and the domain-of-validity test, the invariant-enforcement certificate and signal-flow quantification enabled by our formal framework yield a principled measure of how faithfully a neural model embodies the physical law it appears to predict — closing a measurement gap that no current framework addresses for laws over continuous variables.

Our contributions are threefold:
\vspace*{-3mm}
\begin{enumerate}
\item  We propose a physics-grounded interpretability framework, LAWFUL, that aligns a learned model with an explicit physics law and witnesses their agreement over possibly continuous counterfactual physical states in the region of validity of the law. 
\vspace*{-2mm}
\item LAWFUL further defines a physical consistency score over a family of counterfactual perturbations and identifies physically consistent circuits as 
subgraphs that recover the score under activation patching.
\vspace*{-2mm}
\item Instantiating LAWFUL on a MoCap-to-Radar Doppler task, we characterize its consistency over the region of validity with respect to the Doppler Frequency Law and its invariance to tangential velocity, and we identify a circuit that recovers \(91\%\) of the model's Doppler consistency using half of its components, with the response to velocity scaling carried by temporal attention patterns rather than value transformation, suggesting an adaptive form of temporal differentiation.
\end{enumerate}

\newcommand{\phys}{g}
\newcommand{\learned}{f}
\newcommand{\patch}[2]{\learned^{#1 \leftarrow #2}}  

\section{Law-Aligned Interpretability Framework}
\label{sec:picc}

Our framework formally defines a measure of how consistently a model behaves with respect to a known physical law over continuous variables based on families of counterfactual states, as well as an attribution of that consistency to subsets of internal components in the model. The formalization is in terms of two concepts: \emph{physical consistency}, a score that measures how closely the model's behavior agrees with the physical law under a family of controlled perturbations (\S\ref{sec:picc:metric}); and \emph{physically consistent circuit}, which is a  subgraph of \(\learned\) that recovers this score under activation patching with the family of counterfactual perturbations (\S\ref{sec:picc:circuit}).

\subsection{Aligning a Neural Network and a Physical Law}
\label{sec:picc:setup}

Our framework pairs a \emph{learned model} \(\learned\) with a \emph{physics model} \(\phys\) that captures a known physical law expected to hold in the domain of \(\learned\). The learned model \(\learned : X \to Y_\learned\) maps inputs from a space \(X\) to outputs in a space \(Y_\learned\); the physics model \(\phys : \Omega \to Y_\phys\) is a function from domain-interpretable quantities in a continuous space \(\Omega\) to outputs in a space \(Y_\phys\). 
To measure how \(\learned\) aligns with \(\phys\) we evaluate the former's behavior against that of the latter. Throughout, we fix an evaluation set \(\{x^{(i)}\}_{i=1}^{N} \subseteq X\) of \(N\) inputs for which consistency is measured.

\subsection{Physical consistency}
\label{sec:picc:metric}
To compare \(\learned\) and \(\phys\), we introduce domain-grounded bridges that express each model in terms of a common physical observable, together with a perturbation family on \(\Omega\).

\vspace*{-2mm}
\paragraph{Input bridge.}
Raw inputs of \(\learned\) can carry a mixture of information, including the physical quantities on which \(\phys\) is defined. We extract these quantities via a map \(B : X \to \Omega\) that carries raw inputs into the physics space, taking \(\Omega := B(X)\) as the physics space reachable by inputs to \(\learned\). The choice of \(B\) determines the physical variables of interest.

\vspace*{-2mm}
\paragraph{Output bridge.}
The two models' outputs can live in different spaces and carry different information densities. We choose a common physical observable \(Z\) and project each model onto it via \(D_\learned : Y_\learned \to Z\) and \(D_\phys : Y_\phys \to Z\). The choice of \(Z\) determines which aspect of the physics is being tested.

\vspace*{-2mm}
\paragraph{Perturbation family.}
We specify a family \(\mathcal{P} = \{p_\alpha\}_{\alpha \in A}\) of physically meaningful perturbation operations \(p_\alpha : \Omega \to \Omega\) that generate counterfactual physical states at which \(\learned\) and \(\phys\) can be compared. For each \((i, \alpha)\), we assume access to \(x'^{(i,\alpha)} \in X\) with \(B(x'^{(i,\alpha)}) = p_\alpha(B(x^{(i)}))\), so that \(\learned\) and \(\phys\) can be evaluated at the same perturbed physical state. 

{\em Remark.}~~The choice of \(\mathcal{P}\) determines which physical relationships in \(\phys\) the analysis probes, and is dictated by the law under study. The choice of $A$ determines the coverage and domain-of-validation in the continuous space \(\Omega\). Computing $\mathcal{P}$ for (countably or uncountably) infinite $A$ involves computing $p_\alpha$ for a finite, potentially sparse set of $\alpha$ samples, and then using approximation, interpolation, or signal reconstruction techniques to span all $\alpha \in A$.

\vspace*{-2mm}
\paragraph{Consistency score.}
We measure the agreement between the bridged outputs of \(\learned\) and \(\phys\) by a scalar functional \(\Phi\). The bridged outputs, for every \(i\) and \(\alpha \in A\), are
\begin{equation}
\begin{aligned}
\bar{z}_\learned^{(i)} &= D_\learned(\learned(x^{(i)})),
  & z_\learned^{(i,\alpha)} &= D_\learned(\learned(x'^{(i,\alpha)})), \\
\bar{z}_\phys^{(i)} &= D_\phys(\phys(B(x^{(i)}))),
  & z_\phys^{(i,\alpha)} &= D_\phys(\phys(B(x'^{(i,\alpha)}))).
\end{aligned}
\label{eq:bridged-outputs}
\end{equation}
The baselines \(\bar{z}_\learned^{(i)}, \bar{z}_\phys^{(i)}\) anchor each model, while the perturbed outputs \(z_\learned^{(i,\alpha)}, z_\phys^{(i,\alpha)}\) reveal how each model responds to \(\mathcal{P}\). Collecting these into \(\mathbf{z}_\learned = \{\bar{z}_\learned^{(i)}, z_\learned^{(i,\alpha)}\}_{i,\alpha}\) and \(\mathbf{z}_\phys = \{\bar{z}_\phys^{(i)}, z_\phys^{(i,\alpha)}\}_{i,\alpha}\), \(\Phi\) takes the form
\[
  \Phi: (\mathbf{z}_\learned, \mathbf{z}_\phys)
  \mapsto \mathbb{R}.
\]
\(\Phi\) is a design choice—pointwise error metrics give absolute deviation, similarity-based measures give response alignment—and we adopt the convention that larger \(\Phi\) indicates closer agreement, with \(\Phi\) at its supremum when $f$ matches $g$'s ideal response. Since \(\mathbf{z}_\learned, \mathbf{z}_\phys\) are determined by \((\learned, \phys)\) once the evaluation set, bridges, and perturbation family are fixed, we define the \emph{consistency score}
\vspace*{-1mm}
\begin{equation}
  M_\mathcal{P}(\learned, \phys)
  \coloneqq
  \Phi(\mathbf{z}_\learned, \mathbf{z}_\phys).
  \label{eq:consistency-score}
\vspace*{-2mm}
\end{equation}

Across choices of \(\Phi\), \(M_\mathcal{P}(\learned, \phys)\) remains physics law-grounded: a change in \(M_\mathcal{P}\) reflects a change in \(\learned\)'s \emph{physically consistent behavior}, i.e., its agreement with \(\phys\) on counterfactual physical states generated by \(\mathcal{P}\).

\subsection{Physically consistent circuit}
\label{sec:picc:circuit}
We now identify which components of \(\learned\) account for the agreement captured by \(M_\mathcal{P}(\learned, \phys)\). Treating \(\learned\) as a computational graph \(G = (V, E)\), where vertices \(V\) are model components (e.g., attention heads, sublayers) and directed edges \(E\) represent the residual-stream information flow between them~\citep{elhage2021mathematical}, we probe subgraphs \(C\) of \(G\) and demonstrate the framework using activation patching.

\vspace*{-2mm}
\paragraph {Configurations for counterfactual perturbations.}
~Let \(\patch{C}{s}(b)\) denote a forward pass on base \(b\) in which the activations on \(C\) are replaced by those captured during a forward pass on source \(s\), following standard activation patching~\citep{wang2023ioi,heimersheim2024activation}. For each \((x^{(i)}, x'^{(i,\alpha)})\), we apply two configurations, both rooted at the unperturbed base \(x^{(i)}\) with the perturbed counterfactual input \(x'^{(i,\alpha)}\) as the activation source. The \emph{sufficiency} configuration \(\patch{C}{x'^{(i,\alpha)}}(x^{(i)})\) routes the perturbation through \(C\) alone, holding the complement \(V\setminus C\) at its base activations. The \emph{necessity} configuration \(\patch{V\setminus C}{x'^{(i,\alpha)}}(x^{(i)})\) routes the perturbation through the complement, holding \(C\) at its base activations. Sufficiency asks whether routing the perturbation through \(C\) alone reproduces \(\learned\)'s response to the full perturbation; necessity asks whether that response survives when the perturbation reaches every component except \(C\). 

\paragraph{Evaluating \(M\) under patching.}
Both configurations modify only the \(\learned\)-side perturbed-state slot \(\{z_\learned^{(i,\alpha)}\}\); the baselines \(\{\bar{z}_\learned^{(i)}\}\) and the physics-side outputs \(\{\bar{z}_\phys^{(i)}, z_\phys^{(i,\alpha)}\}\) remain unchanged. For the sufficiency configuration, \(z_\learned^{(i,\alpha)}\) is replaced by \(D_\learned(\patch{C}{x'^{(i,\alpha)}}(x^{(i)}))\); for the necessity configuration, by \(D_\learned(\patch{V\setminus C}{x'^{(i,\alpha)}}(x^{(i)}))\). Substituting into Eq.~\eqref{eq:consistency-score} yields the patched consistency scores \(M_\mathcal{P}^{\mathrm{suf}}(C, \phys)\) and \(M_\mathcal{P}^{\mathrm{nec}}( C, \phys)\), evaluated against the unpatched baseline \(M_0 \coloneqq M_\mathcal{P}(\learned, \phys)\).

\paragraph{Interpreting the patched scores.}
We measure each patched score relative to the unpatched baseline score, \(M_0\), by
\vspace*{-2mm}
\begin{align}
  \Delta_{\mathrm{suf}}(C) &\coloneqq M_0
    - M_\mathcal{P}^{\mathrm{suf}}(C, \phys), \nonumber \\
  \Delta_{\mathrm{nec}}(C) &\coloneqq M_0
    - M_\mathcal{P}^{\mathrm{nec}}(C, \phys).
  \label{eq:delta-suf-nec}
\vspace*{-2mm}
\end{align}

A sufficiency deviation \(\Delta_{\mathrm{suf}}(C)\)
close to zero indicates that routing the perturbation
through \(C\) alone recovers \(\learned\)'s physical
consistency. A large necessity deviation
\(\Delta_{\mathrm{nec}}(C)\) indicates that routing the
perturbation through every component except \(C\) fails
to recover it. 

\vspace*{-2mm}
\paragraph{Definition of physically consistent circuit.}
\label{def:pcc}
For \(\tau \in (0, 1)\) and \(M_0 > 0\), a subgraph
\(C\) is \emph{\(\tau\)-consistent} for
\((\learned, \phys, \mathcal{P})\) if
\begin{align*}
  \Delta_{\mathrm{suf}}(C) &\le (1 - \tau)\, M_0, \\
  \Delta_{\mathrm{nec}}(C) &\ge \tau\, M_0.
\end{align*}
A \(\tau\)-consistent \(C\) is \emph{minimal} if no
\(C \setminus \{c\}\) for \(c \in C\) is itself
\(\tau\)-consistent. A \emph{\(\tau\)-physically
consistent circuit} is a minimal \(\tau\)-consistent
subgraph. 

{\em Remark.}~~~The patching targets recovery of the unpatched baseline \(M_0\), so the resulting analysis is specific to physically consistent behavior: a subgraph \(C\) that recovers \(M_0\) accounts for \(\learned\)'s agreement with \(\phys\) under \(\mathcal{P}\). This specializes the standard notion of circuit by taking the recovery target to be physical consistency rather than task output, and the identified circuit therefore depends on \(\mathcal{P}\): different perturbation families may expose different subgraphs of \(\learned\).

\section{Instantiation: Doppler Frequency Consistency in MoCap-to-Radar}
\label{sec:inst}
We instantiate the framework for the physical-consistency evaluation of \citet{chen2026physics}: we specify the learned transformer model \(\learned\) and physics model \(\phys\) (\S\ref{sec:inst:setup}); the bridges, perturbation families, and consistency scores (\S\ref{sec:inst:metric}); and the circuit discovery procedure on the resulting computational graph (\S\ref{sec:inst:circuit}).

\subsection{Transformer and Doppler frequency model}
\label{sec:inst:setup}
We pair a MoCap-to-radar transformer \(\learned\) with a Doppler frequency model \(\phys\), both instantiated on a single continuous MoCap recording of 3D marker positions, with \(x_{m,t} \in \mathbb{R}^3\) denoting the 3D position of marker \(m\) at sample index \(t\), with \(m=1,\ldots,M\). The recording is partitioned into \(N\) overlapping STFT windows of length \(L\); each window \(x^{(i)} \in \mathbb{R}^{M \times L \times 3}\) is fed to \(\learned\), forming the evaluation set \(\{x^{(i)}\}_{i=1}^N\) for which consistency is measured. We suppress the window index \((i)\) henceforth, except when aggregating across windows.

The learned model \(\learned\) is a spatio-temporal transformer that composes a
spatial transformer over markers with a temporal transformer over frames, each a
single self-attention layer followed by an MLP, to predict a dB-scale
micro-Doppler spectrum \(S^{\mathrm{pred}} = \learned(x) \in \mathbb{R}^{K}\),
where \(K\) is the number of Doppler-frequency bins.


The physics model \(\phys\) applies the Doppler frequency law independently to each MoCap marker:
\vspace*{-2mm}
\begin{equation}
  \phys(v^{\mathrm{rad}})
    \;=\; \frac{2\, v^{\mathrm{rad}}}{\lambda},
  \label{eq:doppler}
\end{equation}
where \(\lambda\) is the radar wavelength. Its linearity implies that scaling every marker's radial velocity by a common \(\alpha\) scales every marker's Doppler frequency by \(\alpha\).


\subsection{Doppler consistency}
\label{sec:inst:metric}

\paragraph{Input bridge.}
Let \(x_{\text{radar}} \in \mathbb{R}^3\) denote the radar location in the MoCap coordinate frame. For each marker \(m\), the range to the radar is \(r_{m,t} = \|x_{m,t} - x_{\text{radar}}\|_{2}\), and radial velocity is approximated by central finite difference on the recording:
\begin{equation}
  v^{\text{rad}}_{m,t} \;\approx\;
  \frac{r_{m,t+1} - r_{m,t-1}}{2\,\Delta t}.
  \label{eq:radial-vel}
\end{equation}
The bridge \(B : \mathbb{R}^{M \times L \times 3} \to \mathbb{R}^{M \times L}\) maps a window \(x\) to the corresponding sequence of \((v^{\text{rad}}_{m,t})\).


\paragraph{Output bridge.}
Both bridges target the Doppler centroid, a scalar summary of how spectral power is distributed across frequency. The learned-side bridge \(D_\learned : \mathbb{R}^{K} \to \mathbb{R}\) computes the power-weighted mean Doppler frequency of the predicted spectrum:
\begin{equation}
  D_\learned\!\left(S^{\mathrm{pred}}\right)
    =
    \frac{\sum_{k=0}^{K-1} P_k\, \nu_k}
         {\sum_{k=0}^{K-1} P_k},
  \label{eq:Dlearned}
\end{equation}
where \(P_k = 10^{S^{\mathrm{pred}}_k/10}\), and
\(\nu_k\) is the Doppler frequency at bin \(k\).

Since radar returns aggregate scattering across the body, the physics-side bridge \(D_\phys : \mathbb{R}^{M \times L} \to \mathbb{R}\) aggregates per-marker Doppler frequencies into a body-level centroid with weights \(w_m\) approximating each marker's radar cross-section (RCS) via body-surface-area (BSA) proportions: 
\begin{equation}
D_\phys(\phys(v^{\mathrm{rad}}))
    \;=\; \frac{1}{L}\sum_{t=1}^{L}
           \sum_{m=1}^{M} \frac{w_m}{\sum_j w_j}\, \phys(v^{\mathrm{rad}}_{m,t}).
  \label{eq:Dphys}
\end{equation}

\paragraph{Perturbation family.}

We perturb radial velocity by uniform scaling: for \(\alpha \in \mathbb{R}\), \(p_\alpha\) scales all radial velocities by \(\alpha\). Since \(v^{\mathrm{rad}}_{m,t}\) is obtained by finite differencing the range, and the first-order change in range equals the line-of-sight component of the frame-to-frame displacement, scaling the radial velocity by \(\alpha\) corresponds to scaling the radial component of the displacement, while leaving the tangential component unchanged. We decompose \(\Delta x_{m,t} = \Delta x^{\text{rad}}_{m,t} + \Delta x^{\text{tan}}_{m,t}\), where \(\Delta x^{\text{rad}}_{m,t} = (\Delta x_{m,t} \cdot \hat r_{m,t-1})\, \hat r_{m,t-1}\) is the projection of \(\Delta x_{m,t}\) onto the line-of-sight direction \(\hat r_{m,t-1} = (x_{m,t-1} - x_{\mathrm{radar}}) / \lVert x_{m,t-1} - x_{\mathrm{radar}} \rVert\) at \(x_{m,t-1}\).  The counterfactual \(x'^{(\alpha)}\) is then generated by
\begin{equation}
\begin{aligned}
  x'^{(\alpha)}_{m,1} &= x_{m,1}, \\
  x'^{(\alpha)}_{m,t} &= x'^{(\alpha)}_{m,t-1}
    + \alpha\, \Delta x^{\text{rad}}_{m,t}
    + \Delta x^{\text{tan}}_{m,t},
\end{aligned}
\label{eq:lift}
\end{equation}
attenuating or amplifying \(\Delta x^{\mathrm{rad}}_{m,t}\) while preserving tangential motion, and realizing \(B(x'^{(\alpha)}) \approx \alpha\, B(x)\).

\paragraph{Consistency score.}

For counterfactuals \(x'^{(\alpha)}\) constructed by scaling radial velocity by \(\alpha\), physical consistency demands that \(\learned\)'s response track \(\phys\)'s across the perturbation family \(\mathcal{P} = \{p_\alpha\}_{\alpha \in A}\), indexed by a set of scaling factors \(A \subset \mathbb{R}\). We summarize each model's response at \(\alpha\) by the least-squares slope of the perturbed bridged output against the unperturbed baseline over the evaluation set,
\vspace*{-2mm}
\begin{equation}
  \hat m_\learned(\alpha)
    \;=\; \frac{\sum_{i=1}^{N} \bar z_\learned^{(i)}\,
                                 z_\learned^{(i,\alpha)}}
               {\sum_{i=1}^{N} \bigl(\bar z_\learned^{(i)}\bigr)^{2}}.
  \label{eq:m-hat}
\end{equation}
Replacing \(\learned\) with \(\phys\) in Eq.~\eqref{eq:m-hat} yields the physics-side slope \(\hat m_\phys(\alpha)\). The
consistency score quantifies how closely \(\hat m_\learned(\alpha)\) matches \(\hat m_\phys(\alpha)\) across \(\mathcal{P}\).

We aggregate the squared deviations between \(\hat m_\learned\) and \(\hat m_\phys\) across \(A\):
\begin{equation}
  M_{\mathcal{P}}(\learned, \phys)
    \;=\; 1 \;-\;
      \frac{\sum_{\alpha \in A}\bigl(\hat m_\learned(\alpha)
                                     - \hat m_\phys(\alpha)\bigr)^{2}}
           {\sum_{\alpha \in A}\bigl(\hat m_\phys(\alpha)
                                     - \bar m_\phys\bigr)^{2}},
  \label{eq:dcs}
\end{equation}
where \(\bar m_\phys = \tfrac{1}{|A|}\sum_{\alpha \in A}
\hat m_\phys(\alpha)\). \(M_{\mathcal{P}} \le 1\), with \(1\)
attained when \(\hat m_\learned\) matches \(\hat m_\phys\)
across \(\mathcal{P}\).

\subsection{Doppler consistency circuit}
\label{sec:inst:circuit}
We identify circuits in $\learned$ following the $\tau$-physical consistency circuit definition of \S\ref{sec:picc:circuit}, beginning with the computational graph $G = (V, E)$ on the spatio-temporal transformer. We analyze the model at component granularity: vertices $V$ comprise the attention heads and MLPs of the spatial transformer and those of the temporal transformer; edges $E$ trace the residual-stream flow within each transformer together with the linear projection bridging the spatial and temporal streams. As a first instantiation aimed at demonstrating the framework, this granularity supports circuit identification at the level of specific heads and MLPs and is well-matched to the modest component count of \(\learned\) in this study.
\section{Physics-Based Interpretation of the MoCap-to-Radar Transformer}
\label{sec:setup}

\begin{figure}[t]
  \centering
  \includegraphics[width=\linewidth]{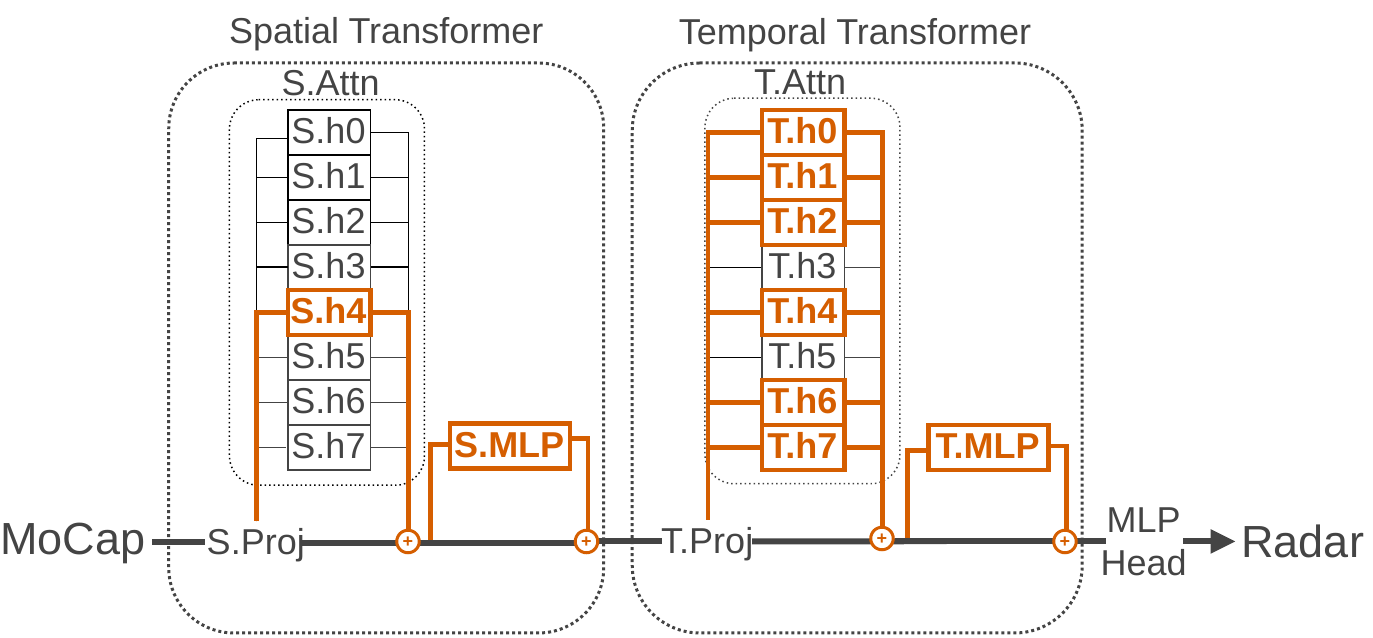}
    \caption{The architecture of the $\mathrm{ST(Flat)}$ spatio-temporal transformer. Components highlighted in
    red form an example of an identified $\tau=0.9$
    physical consistency circuit, as reported in
    Table~\ref{tab:circuits}.}
    \label{fig:arch}
\end{figure}

We now identify circuits in $\learned$ and analyze their physical consistency with the Doppler model $\phys$. We follow the experimental setup of \citet{chen2026physics}, using their $\mathrm{ST(Flat)}$ MoCap-to-Radar architecture which exposes $|V| = 18$ patchable components (Figure~\ref{fig:arch}). We use the $\mathrm{RandomWalk}$ dataset as our primary evaluation set. 

\vspace*{-2mm}
\paragraph{Domain of validity for perturbation family.}
Ideally, consistency scoring and circuit identification should use continuous perturbations within the radial velocity bounds that the radar can resolve. The Bumblebee radar used in this study resolves radial velocities only up to 2.6\,m/s \citep{samraksh_bumblebee_product}. Using the same RCS weights \(w_m\) as in Eq.~\eqref{eq:Dphys}, the maximum training-set body-level radial-velocity magnitude is 2.24\,m/s. We accordingly estimate the \emph{radar-resolvable} scaling regime to be \(\alpha \in A = [-\alpha_\text{max}, \alpha_\text{max}]\) with \(\alpha_\text{max} = 2.6/2.24 \approx 1.16\), within which every frame's body-level radial velocity stays below the radar velocity limit; $\alpha$ outside this range exceeds the radar's resolvable velocity. 

\subsection{Physically consistent circuit identification}

\paragraph{Unpatched consistency baseline.}
The consistency score of the unpatched model is $M_0 = 0.988$, reflecting near-perfect agreement with $\phys$ before any patching. This baseline establishes that physical consistency is present in $\learned$, motivates the search for specific components that carry it, and defines the reference against which patched scores are measured: subgraph contributions are quantified by the degradation they induce relative to $M_0$.

\vspace*{-2mm}
\paragraph{Incremental search procedure.}
We construct a family $\{C(\tau)\}_\tau$ of $\tau$-consistent circuits by greedy bottom-up search, processing $\tau$ in increasing order. For each $\tau$, we initialize $C$ from the previous circuit (or $\emptyset$ for the smallest $\tau$), iteratively add the component $c$ that most reduces $\Delta_{\mathrm{suf}}(C \cup \{c\})$ until $C$ is $\tau$-consistent, then prune the components added at the current $\tau$ to satisfy minimality. By construction, $C(\tau_i) \subseteq C(\tau_j)$ whenever $\tau_i < \tau_j$, and each $C(\tau)$ is minimal among $\tau$-consistent supersets of its predecessor. We sweep $\tau \in \{0.3, 0.4, \dots, 0.9\}$. 

\vspace*{-2mm}
\paragraph{Controls.}
We compare \(C(\tau)\) against a \emph{source-randomized control}, applied in both patching configurations. In the sufficiency configuration, the control patches the components in \(C(\tau)\) but draws source activations from a random evaluation window \(x'^{(j,\alpha)}\), with \(j \ne i\), independently for each \((i,\alpha)\) pair, yielding \(\Delta^{\mathrm{rand}}_{\mathrm{suf}}\). In the necessity configuration, the control patches the components in \(V \setminus C(\tau)\) and draws source activations from random evaluation windows in the same manner, yielding \(\Delta^{\mathrm{rand}}_{\mathrm{nec}}\). These controls test whether the effect of patching depends on the matched counterfactual source, rather than only on patching the same locations. Reported \(\pm\) values denote standard errors over ten random seeds, with each seed independently sampling \(j \! \ne \! i\) for every \((i,\alpha)\) pair.

\vspace*{-2mm}
\paragraph{Search results.}

\begin{table*}[t]
\centering
\caption{Reported  \(\tau\)-physical consistency
circuits. Component notation:
T / S denote temporal / spatial transformer branches; h\(k\) the
\(k\)-th attention head; MLP the feed-forward block.
Source-randomized controls
\(\Delta^{\mathrm{rand}}_{\mathrm{suf}}\) and
\(\Delta^{\mathrm{rand}}_{\mathrm{nec}}\) patch the
same components as \(C\) but draw source activations
from a random evaluation window.}
\label{tab:circuits}
\begin{tabular}{@{}c c l c c c c@{}}
\toprule
$|C|$ & $\tau$ & circuit $C$ &
$\Delta_{\mathrm{suf}}$  &
$\Delta_{\mathrm{nec}}$ &
$\Delta^{\mathrm{rand}}_{\mathrm{suf}}$ &
$\Delta^{\mathrm{rand}}_{\mathrm{nec}}$  \\
\midrule
3 & 0.3--0.5 & T:\{MLP, h6, h7\} & 0.49 & 1.89 & $1.82 \pm 0.04$ & $1.11 \pm 0.02$ \\
\addlinespace[6pt]
4 & 0.6 & T:\{MLP, h1, h6, h7\} & 0.35 & 2.43 & $2.11 \pm 0.04$ & $1.29 \pm 0.02$ \\
\addlinespace[6pt]
5 & 0.7 & T:\{MLP, h1, h6, h7\}, S:\{MLP\} & 0.24 & 3.08 & $2.39 \pm 0.04$ & $2.17 \pm 0.02$ \\
\addlinespace[6pt]
6 & 0.8 & \makecell[l]{T:\{MLP, h1, h4, h6, h7\},S:\{MLP\}} & 0.18 & 3.19 & $2.52 \pm 0.04$ & $2.03 \pm 0.02$\\
\addlinespace[6pt]
9 & 0.9 & \makecell[l]{T:\{MLP, h0, h1, h2, h4, h6, h7\}, S:\{MLP, h4\}} & 0.09 & 3.50 & $3.05 \pm 0.05$ & $3.23 \pm 0.02$ \\
\bottomrule
\end{tabular}
\end{table*}

Doppler consistency is concentrated in a temporal-block core. As shown in Table~\ref{tab:circuits}, the smallest \(\tau\)-consistent circuit (\(|C|=3\)) consists of three temporal-block components,
T:\{MLP, h6, h7\}, and circuits at higher \(\tau\) expand almost entirely within the temporal block; the spatial block contributes only marginally, with S:\{MLP\} entering at \(\tau=0.7\) and an additional
spatial head only at \(\tau=0.9\). The source-randomized controls satisfy  
\(\Delta^{\mathrm{rand}}_{\mathrm{suf}}
> \Delta_{\mathrm{suf}}\) and
\(\Delta^{\mathrm{rand}}_{\mathrm{nec}}
< \Delta_{\mathrm{nec}}\) at every \(|C|\), 
so this concentration is specific to the matched counterfactual source, not an artifact of the patched locations. For the analysis and discussion that follows, we anchor on the
\emph{primary circuit}
\[
C^\dagger = T: \{\mathrm{MLP}, h_0, h_1, h_2, h_4, h_6, h_7\},\ 
S: \{\mathrm{MLP}, h_4\}, 
\]
which, at \(\tau = 0.9\), recovers 91\% of \(M_0\) 
using only half (\(|C^\dagger|/|V| = 9/18\)) of the components, most of which
lie in the temporal block (7 out of 9).


\vspace*{-2mm}
\paragraph{Visualizing physical consistency of \(C^\dagger\).}
\label{sec:res:identify}
\begin{figure*}[t]
  \centering
  \includegraphics[width=\textwidth]{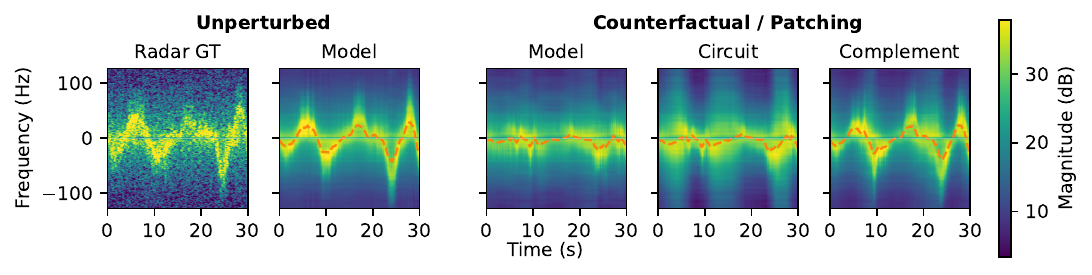}
\caption{GT, predicted, and patched Doppler spectrograms for a representative 30s evaluation clip with perturbation \(\alpha = 0.3\). Orange dashed curves denote Doppler centroids computed by the output bridge \(D_\learned\) in Eq.~\eqref{eq:Dlearned}. The left \textit{Unperturbed} group contains the ground-truth radar spectrogram (\textit{Radar GT}) and the model prediction \(\learned(x)\) (\textit{Model}). The right \textit{Counterfactual} group contains the model prediction \(\learned(x')\) (\textit{Model}) and two patching-based outputs: the sufficiency configuration \(\patch{C^\dagger}{x'}(x)\) (\textit{Circuit}) and the necessity configuration \(\patch{V\setminus C^\dagger}{x'}(x)\) (\textit{Complement}).}
\label{fig:circuit-vis}
\end{figure*}

The primary circuit $C^\dagger$ reproduces the physics-consistent Doppler centroid; its complement does not. As shown in Figure~\ref{fig:circuit-vis}, the Counterfactual Model panel shows $\learned$'s Doppler centroid compressed in amplitude relative to Unperturbed. This {\em compression is consistent with the $\alpha = 0.3$ scaling} applied to the radial velocities of $x'$; the Counterfactual Circuit panel reproduces this compressed trajectory, while the Counterfactual Complement panel {\em retains the original amplitude} seen in the Unperturbed Model panel.

\subsection{Counterfactual behavior over domain of validity}
\label{sec:res:eval}
\begin{figure}[t]
    \centering
    \begin{subfigure}[t]{0.33\textwidth}
        \hspace*{-22pt}%
        \includegraphics[width=\linewidth]{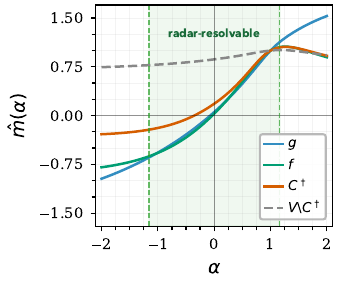}
        \caption{}
        \label{fig:physics-eval-rad}
    \end{subfigure}
    \hfill
    \begin{subfigure}[t]{0.33\textwidth}
        \hspace*{-20pt}%
        \includegraphics[width=\linewidth]{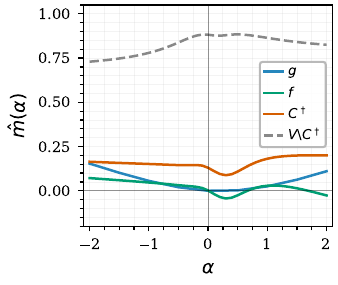}
        \caption{}
        \label{fig:physics-eval-tan}
    \end{subfigure}
    \caption{Doppler frequency consistency under counterfactual velocity perturbations, comparing $f$ (learned model), $g$ (physics model), $C^\dagger$ (Circuit), and $V \setminus C^\dagger$ (Complement). Each panel plots the slope $\hat{m}(\alpha)$ versus scaling factor $\alpha$.
    \textbf{(a)} Radial velocity perturbations, shaded band indicating the domain of validity $|\alpha| \le \alpha_\text{max} \approx 1.16$.
    \textbf{(b)} Tangential velocity perturbations, illustrating invariance.}
    \label{fig:physics-eval}
\end{figure}

We evaluate both the learned model $\learned$ and the circuit $C^{\dagger}$ by comparing their responses to counterfactual physical states against the responses predicted by the physics model $\phys$. The evaluation focuses on two physically meaningful aspects of their behavior: (i) how they behave across the domain of validity, with $\alpha \in [-1.16, 1.16]$; this includes characterizing the response to sign-reversed perturbations that invert the radial motion direction, and then beyond to $\alpha \in [-2, 2]$; 
and (ii) how they behave across perturbations that affect the Doppler frequency (radial) versus perturbations that should not (tangential).



\vspace*{-2mm}
\paragraph{Radial velocity scaling over radar-resolvable domain.}
Both $\learned$ and $C^\dagger$ exhibit the behavior prescribed by the Doppler equation: an approximately linear response to radial scaling within the radar-resolvable range. 
As shown in Fig.~\ref{fig:physics-eval-rad}, the {\em learned model \(\learned\) tracks $\phys$ closely throughout the radar-resolvable domain}, accounting for the high $M_0$ reported earlier. The circuit $C^\dagger$ also tracks $\phys$ across this range and is particularly tight on $\alpha \in [0, 1]$, while the complement $V \setminus C^\dagger$ remains nearly flat and shows little response to scaling. Beyond the radar-resolvable range, both $\learned$ and $C^\dagger$ no longer closely follow $\phys$. This discrepancy should not be interpreted as evidence of a single failure mode: it may arise from physical extrapolation beyond the radar's resolvable velocity range, distributional extrapolation beyond the support of the training data, or an interaction between the two. On the negative-$\alpha$ side, $C^\dagger$ loses tracking earlier, which is suggestive: negative scaling combines magnitude scaling with a sign flip, and components outside the identified circuit may be responsible for the latter. Identifying such a sign-flip substructure would require either a perturbation family that isolates the sign operation or a consistency metric sensitive to it, which we leave to future work.

\vspace*{-2mm}
\paragraph{Tangential velocity scaling as a Doppler-invariance test.}
Tangential motion is orthogonal to the radar line-of-sight and  contributes no first-order Doppler shift. A Doppler-consistent  model is therefore expected to be invariant under tangential scaling: scaling the tangential component of motion should not  alter the predicted radar response. Because Doppler responds  to radial but not tangential motion, the radial counterfactual in Eq.~\eqref{eq:lift} can preserve the tangential component without affecting $\hat{m}(\alpha)$, but the mirror design for the tangential test would leave a Doppler-active radial background and contaminate the slope. We therefore zero the radial displacement, so that $\hat{m}(\alpha)$ reflects the model's response to tangential motion alone: 
\begin{align*}
    x'^{(\alpha)}_{m,t} = x'^{(\alpha)}_{m,t-1}
    + \alpha\,\Delta x^{\text{tan}}_{m,t}.
\end{align*}
As shown in Fig.~\ref{fig:physics-eval-tan}, under tangential scaling, $\learned$, $C^\dagger$, and $V \setminus C^\dagger$ {\em all exhibit small responses,  consistent with the expected near-invariance under tangential scaling}. The point $\alpha = 0$ corresponds to a stationary body and provides the strongest physics constraint on this panel: the radar response should vanish exactly. Both $\learned$ and $\phys$ meet this constraint, while $C^\dagger$ shows a slight deviation and the complement $V \setminus C^\dagger$ deviates substantially.

\subsection{Component-level mechanism}
\label{sec:res:mech}

The circuit identification analysis establishes that $C^\dagger$ recovers physical consistency at the output, but does not characterize what each component contributes internally. Because the input in this case study does not directly encode radial velocity as an explicit feature, we ask: (i) where in $C^\dagger$ does a representation of radial velocity emerge, and how is it propagated to downstream components; and (ii) within these components, what mechanism produces the Doppler consistency response to scaling perturbations?

\begin{table}[t]
\caption{Linear probe $R^2$ for per-marker radial velocity $v^{\text{rad}}$; QK and value patching in sufficiency-configuration evaluated on the attention heads of $C^{\dagger}$.}
\label{tab:probing-circuit}
\centering
\footnotesize
\setlength{\tabcolsep}{3pt}
\renewcommand{\arraystretch}{1.08}
\begin{tabular}{@{}p{0.34\columnwidth}ccc@{}}
\toprule
 & $R^2$ ($\text{per-marker }v^\text{rad}$) & $\Delta_{\mathrm{suf}}^{\mathrm{QK}}$ & $\Delta_{\mathrm{suf}}^{\mathrm{VAL}}$ \\
\midrule
\multicolumn{4}{@{}p{\columnwidth}@{}}{\textit{Non-attention}} \\
T.Proj               & 0.019  & --- & --- \\
T.MLP                & 0.556  & --- & --- \\
T.MLP (Attn ablated) & 0.0004 & --- & --- \\
\midrule
\multicolumn{4}{@{}p{\columnwidth}@{}}{\textit{Attention heads — in $C^{\dagger}$}} \\
T.h0 & 0.078 & 3.06 & 3.48 \\
T.h1 & 0.042 & 2.68 & 3.56 \\
T.h2 & 0.213 & 3.14 & 3.54 \\
T.h4 & 0.057 & 3.45 & 3.28 \\
T.h6 & 0.152 & 1.88 & 3.47 \\
T.h7 & 0.066 & 2.34 & 3.33 \\
S.h4 & 0.032 & 3.55 & 2.92 \\
\midrule
\multicolumn{4}{@{}p{\columnwidth}@{}}{\textit{All attention heads combination — in $C^{\dagger}$}} \\
$C^{\dagger}_{\mathrm{attn}}$ & 0.761 & 0.35 & 1.31 \\
\bottomrule
\end{tabular}
\end{table}

\vspace*{-2mm}
\paragraph{Emergence and propagation of radial-velocity representations.}
We use linear probing methods~\citep{alain2016understanding} to characterize the components of $C^\dagger$ and the projection layer (T.Proj). Table~\ref{tab:probing-circuit} reports linear-probe \(R^2\) scores for decoding per-marker radial velocity, from the activations of components in \(C^\dagger\). The pre-attention temporal activations, T.Proj, contain little linearly decodable radial-velocity information ($R^2 = 0.019$). Likewise, no head individually decodes $v_{\text{rad}}$ above $R^2 = 0.213$. In contrast, their joint representation reaches $R^2 = 0.761$, indicating that $v_{\text{rad}}$ becomes linearly available through the combined temporal-attention output. T.MLP also decodes $v_{\text{rad}}$ ($R^2 = 0.556$), but replacing the attention output with its evaluation-set mean collapses this
to $R^2 = 0.0004$. Together with the per-head and $C^\dagger$-level results, these observations establish radial-velocity emergence in the combined output of $C^\dagger$ temporal heads and its propagation
through T.MLP.

\vspace*{-4mm}
\paragraph{Mechanism within temporal attention.}

We have established that temporal heads dominate Doppler frequency consistency and that radial velocity emerges from their combined output. We now ask what mechanism gives rise to this emergence: is the scaling response carried by attention pattern formation (QK) or value transformation (VAL)? Within our framework, we extend sufficiency patching to these two pathways: for each attention head in $C^\dagger$, we replace either its attention pattern (QK) or its value activations (VAL) with those captured during a counterfactual forward pass on $x'^{(\alpha)}$, holding the complementary pathway at base.

For a sub-circuit $c \subseteq C^\dagger_\text{attn}$, where $C^\dagger_\text{attn}$ denotes all the attention heads in $C^\dagger$, we define
\begin{align*} 
  \Delta_{\mathrm{suf}}^{\mathrm{QK}}(c) &\coloneqq M_0 - M_\mathcal{P}^{\mathrm{QK\text{-}suf}}(c, \phys), \\
  \Delta_{\mathrm{suf}}^{\mathrm{VAL}}(c) &\coloneqq M_0 - M_\mathcal{P}^{\mathrm{VAL\text{-}suf}}(c, \phys),
\end{align*}
in direct analogy to Eq.~\eqref{eq:delta-suf-nec}. We use the same perturbation family $\mathcal{P}$ as in circuit identification.

As shown in Table~\ref{tab:probing-circuit}, $\Delta_{\mathrm{suf}}^{\mathrm{QK}}(C^\dagger_{\mathrm{attn}})$ is substantially smaller than $\Delta_{\mathrm{suf}}^{\mathrm{VAL}}(C^\dagger_{\mathrm{attn}})$ (0.35 vs.\ 1.31), indicating that replacing the attention pattern alone is sufficient to recover physical consistency, whereas replacing value activations alone is not. At the single-head level, six of the seven temporal heads in $C^\dagger$ are more QK-sufficient than VAL-sufficient (the only exception is T.h4). The spatial head S.h4 is the only head where VAL is more sufficient than QK, but S.h4 is also the least QK-sufficient ($\Delta_\text{suf}^\text{QK} = 3.55$ is the largest deviation in the table, indicating the weakest QK contribution), so its VAL contribution is marginal. Within the temporal block, T.h1, T.h6, and T.h7 are the three most QK-sufficient heads, and they coincide with the heads that saturate sufficiency earliest in circuit identification (Table~\ref{tab:circuits}). Taken together, the QK contribution is concentrated and structurally aligned  with the circuit's sufficiency ordering, while the VAL contribution is diffuse and small.


We interpret this as evidence that $\learned$ adaptively computes temporal differences across frames. Radial velocity is recovered by comparing range measurements between frames, and the most useful time gap depends on how fast the target is moving. Short gaps capture fast motion, whereas longer gaps are more suitable for slow motion. The QK-dominant and temporally localized attribution suggests that the attention pattern controls this choice by selecting the frame pairs used for the comparison. This is similar to multi-scale temporal differentiation in classical signal processing.

\section{Limitations}
\label{sec:limits}
 
We instantiate LAWFUL on a single case study: one linear law (the
Doppler relation) in a spatio-temporal transformer. Nonlinear or
multi-variable laws, and larger models, may require different bridges,
perturbation families, and search procedures than the ones we use here,
and characterizing the bridged outputs at scale remains open. LAWFUL
also takes the grounding law as given, and does not address how
it should be chosen---among candidate laws or levels of
description---or whether it could be discovered rather than supplied.
 
A subtler limit concerns coverage of the domain of validity. LAWFUL's
perturbations generate out-of-distribution physical states, and we
observe that the model continues to satisfy the Doppler law across
them---evidence of physical generalization beyond the training
distribution. These perturbations rescale the speed of recorded
motions uniformly across the body. Perturbations that change the motion
itself, or that rescale only part of the body, remain untested. Finally, our output bridge reduces each spectrogram to
its Doppler centroid, so the analysis characterizes the bulk shift of
the distribution rather than its finer structure---spread, harmonics,
micro-Doppler---and circuits carrying those features need not coincide
with the one we identify.

\section{Conclusion}
\label{sec:conclusion}

This paper develops physics-aware mechanistic interpretability: in its LAWFUL framework a known physical law, rather than task accuracy or language-model-shaped benchmarks, supplies the ground truth against which a model's internal computation is evaluated. 
We expect the combination of physics-grounded perturbations and mechanism-level attribution to extend to other physics-based interpretability settings, especially for the emerging world models of physical AI.

\bibliographystyle{icml2025}
\bibliography{ref}
\newpage
\appendix
\onecolumn
\section{Notation Summary}
\label{app:notation}
Table~\ref{tab:notation} summarizes the notation used throughout the paper.
\begin{center}
\captionof{table}{Summary of notation.}
\label{tab:notation}
\small
\begin{tabular}{@{}ll@{}}
\toprule
Symbol & Description \\
\midrule
\multicolumn{2}{@{}l}{\textit{LAWFUL framework}} \\
\addlinespace[2pt]
$f : X \to Y_f$ & learned model, from input space $X$ to output space $Y_f$ \\
$g : \Omega \to Y_g$ & physics model, from continuous physics space $\Omega$ to output space $Y_g$ \\
$B : X \to \Omega$ & input bridge; $\Omega := B(X)$ is the physics space reachable by inputs to $f$ \\
$Z$ & common physical observable space \\
$D_f : Y_f \to Z$,\; $D_g : Y_g \to Z$ & output bridges projecting the learned and physics models onto $Z$ \\
$\{x^{(i)}\}_{i=1}^{N}$ & evaluation set of $N$ inputs on which consistency is measured \\
$\mathcal{P} = \{p_\alpha\}$ & perturbation family of physically meaningful operations $p_\alpha : \Omega \to \Omega$ \\
$\alpha$ & perturbation index\\
$x'^{(i,\alpha)}$ & counterfactual input satisfying $B(x'^{(i,\alpha)}) = p_\alpha(B(x^{(i)}))$ \\
$\bar z^{(i)}_f,\; z^{(i,\alpha)}_f$ & bridged learned-model outputs: unperturbed baseline and perturbed response \\
$\bar z^{(i)}_g,\; z^{(i,\alpha)}_g$ & bridged physics-model outputs: unperturbed baseline and perturbed response \\
$\Phi$ & scalar agreement functional on the collected bridged outputs; larger indicates closer agreement \\
$M_{\mathcal{P}}(f,g)$ & physical consistency score, aggregating the agreement measured by $\Phi$ \\
$M_0$ & unpatched baseline score, the value of $M_{\mathcal{P}}(f,g)$ before patching \\
\midrule
\multicolumn{2}{@{}l}{\textit{Physically consistent circuit}} \\
\addlinespace[2pt]
$G = (V,E)$ & computational graph of $f$: components $V$, residual-stream edges $E$ \\
$C \subseteq V$ & candidate circuit, a subgraph of $G$ \\
$c$ & a single component of $f$ \\
$f^{C \leftarrow s}(b)$ & forward pass on base $b$ with activations on $C$ patched from source $s$ \\
$M^{\mathrm{suf}}_{\mathcal{P}}(C,g)$,\; $M^{\mathrm{nec}}_{\mathcal{P}}(C,g)$ & consistency score under sufficiency / necessity patching \\
$\Delta_{\mathrm{suf}}(C)$,\; $\Delta_{\mathrm{nec}}(C)$ & deviations of the patched scores from $M_0$ \\
$\Delta^{\mathrm{rand}}_{\mathrm{suf}}$,\; $\Delta^{\mathrm{rand}}_{\mathrm{nec}}$ & source-randomized control counterparts \\
$\tau$ & consistency threshold defining $\tau$-consistent circuits \\
$C(\tau)$ & circuit identified at threshold $\tau$ by the incremental greedy search \\
$C^\dagger$ & primary circuit for experimental analysis \\
$C^\dagger_{\mathrm{attn}}$ & the attention heads of $C^\dagger$ \\
$\Delta^{\mathrm{QK}}_{\mathrm{suf}}(c)$,\; $\Delta^{\mathrm{VAL}}_{\mathrm{suf}}(c)$ & sufficiency deviation under attention-pattern-only (QK) / value-only (VAL) patching \\
\midrule
\multicolumn{2}{@{}l}{\textit{Doppler instantiation}} \\
\addlinespace[2pt]
$x_{m,t}$ & 3D position of marker $m$ at sample index $t$ \\
$M$ & number of MoCap markers; distinct from the score $M_{\mathcal{P}}$ \\
T.$h_k$, S.$h_k$;\; T.MLP, S.MLP & the $k$-th attention head and the feed-forward block of the temporal (T) / spatial (S) transformer \\
T.Proj, S.Proj;\; T.Out, S.Out & input-projection and output layers of the temporal (T) / spatial (S) block \\
$L$ & STFT window length\\
$\lambda$ & radar wavelength in the Doppler law $f(t) = 2v(t)/\lambda$ \\
$v^{\mathrm{rad}}_{m,t}$ & radial velocity of marker $m$, by central finite difference of the range to the radar \\
$w_m$ & RCS weight of marker $m$ (body-surface-area proportion) \\
$\hat m_f(\alpha)$,\; $\hat m_g(\alpha)$ & least-squares response slopes of $f$ and $g$ at scaling $\alpha$ \\
$\alpha_{\max}$ & radar-resolvable bound, $\alpha \in [-\alpha_{\max}, \alpha_{\max}]$ \\
\bottomrule
\end{tabular}
\end{center}
\clearpage
\section{Training Setup}
\label{app:training}
The MoCap-to-Radar model maps each window of 256 MoCap frames to the
Doppler spectrum of the STFT frame spanning the same interval, with
windows advancing at the STFT's 32-frame stride. Marker coordinates and target spectra are standardized (zero mean, unit variance) using training-set statistics only. The train/validation split is at the recording level (six/two walking recordings); the random-walk recordings used for all counterfactual analyses are held out entirely. Table~\ref{tab:training-hparams} lists the optimization hyperparameters. All experimental results use the lowest-validation-loss checkpoint.

\begin{table}[h]
\caption{Training hyperparameters for the MoCap-to-Radar model.}
\label{tab:training-hparams}
\centering
\begin{tabular}{ll}
\toprule
Parameter count & $3.2$\,M \\
Loss & L1 on standardized log-magnitude spectra \\
Optimizer & AdamW \\
Learning rate & $10^{-4}$ \\
Weight decay & $0.01$ \\
Batch size & 32 \\
LR schedule & halve on validation-loss plateau (patience 10) \\
Early stopping & patience 40 epochs \\
Maximum epochs & 200 \\
\bottomrule
\end{tabular}
\end{table}
\FloatBarrier

\section{Circuit Search Procedure}
\label{app:circuit-search}
Algorithm~\ref{alg:greedy-search} formalizes the greedy bottom-up search
of Section~\ref{sec:setup}. We specify here the pruning setup, the role
of necessity, and the exact minimality it guarantees.
\begin{algorithm}[ht]
\caption{Incremental greedy search with multi-ordering pruning}
\label{alg:greedy-search}
\begin{algorithmic}[1]
\REQUIRE component set $V$; threshold set
  $\mathcal{T} = \{0.3, 0.4, \dots, 0.9\}$;
  baseline score $M_0$; pruning trials $R$
\ENSURE nested family
  $\{C(\tau)\}_{\tau \in \mathcal{T}}$
\STATE $C \gets \emptyset$;\quad $C_{\mathrm{prev}} \gets \emptyset$
\FOR{$\tau \in \mathcal{T}$ in ascending order}
  \WHILE{$\Delta_{\mathrm{suf}}(C) > (1-\tau)\,M_0$
    \AND $C \neq V$}
    \STATE $c^\star \gets
      \arg\min_{c \in V \setminus C}
      \Delta_{\mathrm{suf}}(C \cup \{c\})$
    \STATE $C \gets C \cup \{c^\star\}$
      \COMMENT{ties broken by a fixed component order}
  \ENDWHILE
  \FOR{$r = 1, \dots, R$}
    \STATE $C_r \gets C$
    \STATE draw a random ordering of $C \setminus C_{\mathrm{prev}}$
    \REPEAT
      \STATE remove from $C_r$ the first
        $c \in C_r \setminus C_{\mathrm{prev}}$ in the drawn ordering
        satisfying
        $\Delta_{\mathrm{suf}}(C_r \setminus \{c\})
        \le (1-\tau)\,M_0$, if any
    \UNTIL{no such component exists}
  \ENDFOR
  \STATE $r^\star \gets
    \arg\min_r
    \bigl(|C_r|,\Delta_{\mathrm{suf}}(C_r)\bigr)$
    \COMMENT{lexicographic order}
  \STATE $C \gets C_{r^\star}$
  \STATE $C(\tau) \gets C$;\quad $C_{\mathrm{prev}} \gets C$
\ENDFOR
\end{algorithmic}
\end{algorithm}
\vspace*{-6mm}
\paragraph{Pruning.}
Pruning uses $R = 8$ random deletion orderings under a fixed seed, and
returns the smallest resulting circuit (ties broken by lower
$\Delta_{\mathrm{suf}}$).
\vspace*{-2mm}
\paragraph{Role of necessity.}
Necessity is not part of the search: components are selected and pruned
using $\Delta_{\mathrm{suf}}$ alone. We evaluate
$\Delta_{\mathrm{nec}}(C(\tau))$ post hoc on each returned circuit and
report it in Table~\ref{tab:circuits}.
\vspace*{-2mm}
\paragraph{Minimality.}
Pruning yields circuits that are irredundant relative to the circuit
found at the previous threshold---no single further component is
removable---rather than globally inclusion-minimal.

\section{Linear Probing on Circuit Components}
\label{app:probing}

We probe each component with a ridge regression ($\alpha^{\text{rid}}{=}1$) against the per-marker signed radial velocity (the projection of the marker's 3D velocity onto the radar line-of-sight unit vector). Probe feature dimensions are 128 for spatial-side components, 32 for each individual temporal head, and 256 for T.Proj, T.Out, and T.MLP; combinations concatenate the component features. The corresponding $\Delta_{\text{suf}}$ and $\Delta_{\text{nec}}$ are additionally reported in Table~\ref{tab:probing-appendix}.

\begin{table*}[h!]
\centering
\small
\begin{tabular}{rlrrr}
\toprule
\# & Component / Probe set & $R^2$ ($\text{per-marker }v^\text{rad}$) & $\Delta_{\text{suf}}$ & $\Delta_{\text{nec}}$ \\
\midrule
\multicolumn{5}{l}{\textit{Spatial side}} \\
1 & S.Proj & 0.0104 & ---    & ---    \\
2 & S.Attn      & 0.1200 & 3.679  & 0.106  \\
3 & S.MLP       & 0.0005 & 3.648  & 0.123  \\
4 & S.Out       & 0.0185 & ---    & ---    \\
\midrule
\multicolumn{5}{l}{\textit{Temporal entry}} \\
5 & T.Proj & 0.0188 & ---    & ---    \\
\midrule
\multicolumn{5}{l}{\textit{Temporal heads — in $C^{\dagger}$}} \\
6  & h0 & 0.0784 & 4.543 & 0.114 \\
7  & h1 & 0.0422 & 3.982 & 0.155 \\
8  & h2 & 0.2134 & 4.547 & 0.126 \\
9  & h4 & 0.0567 & 4.452 & 0.120 \\
10 & h6 & 0.1524 & 2.996 & 0.251 \\
11 & h7 & 0.0655 & 3.306 & 0.264 \\
\midrule
\multicolumn{5}{l}{\textit{Temporal heads — not in $C^{\dagger}$}} \\
12 & h3 & 0.0172 & 4.706 & 0.085 \\
13 & h5 & 0.0145 & 4.181 & 0.089 \\
\midrule
\multicolumn{5}{l}{\textit{Circuit combinations}} \\
14 & $C^{\dagger}$ heads only (no MLPs)         & 0.7614 & ---             & ---            \\
15 & full $C^{\dagger}$ (9 components)          & 0.7699 & {0.092}  & {3.503} \\
16 & complement $V \setminus C^{\dagger}$       & 0.3469 & 3.503           & 0.092          \\
\midrule
\multicolumn{5}{l}{\textit{T.MLP}} \\
17 & T.MLP (baseline)                           & 0.5556 & 1.295 & 0.202 \\
\midrule
\multicolumn{5}{l}{\textit{Temporal exit}} \\
18 & T.Out & 0.7072 & ---    & ---    \\
\midrule
\multicolumn{5}{l}{\textit{Causal intervention}} \\
19 & {T.MLP (Attn ablated)} & \textbf{0.0004} & ---    & ---    \\
\bottomrule
\end{tabular}
\caption{Linear probing $R^2$ on RandomWalk eval (mean across 5
time-contiguous CV folds, gap${=}8$ windows) and sufficiency/necessity deviations.}
\label{tab:probing-appendix}
\end{table*}

\clearpage
\section{Temporal Attention Heatmaps}
\begin{figure*}[h!]
\centering
\includegraphics[width=\linewidth]{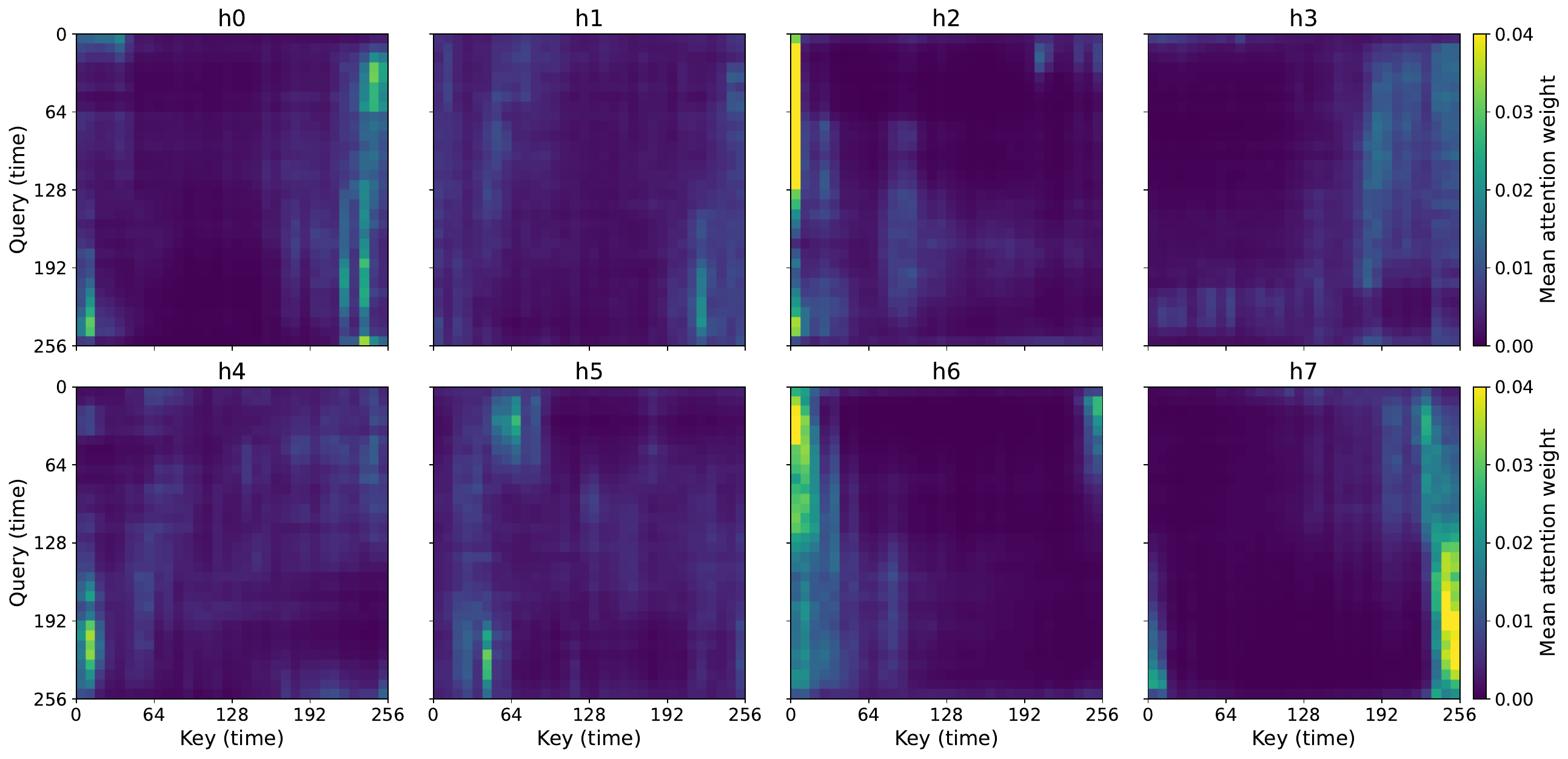}
\caption{Per-head attention patterns of the temporal transformer. Each panel shows the mean attention map of one head ($h_0$--$h_7$), averaged over 200 evaluation windows. Maps are down-sampled from $L \times L = 256 \times 256$ to $32 \times 32$ via $8 \times 8$ block-mean.}
\label{fig:attn_heatmap}
\end{figure*}

We include the temporal attention heatmaps on the held-out random walk recording in Figure~\ref{fig:attn_heatmap}. The heatmaps reveal a complementary
key-selection structure between T.h6 and T.h7: T.h6 concentrates attention
on the earliest frames of the window, most strongly for queries in the
first half, while T.h7 concentrates on the latest frames, most strongly
for queries in the second half. This complementarity could explain why h6 and h7 are the only attention
heads in the smallest $\tau$-consistent circuit, T:\{MLP, h6, h7\}, which
persists across $\tau \in \{0.3, 0.4, 0.5\}$ (Table~\ref{tab:circuits}): a temporal contrast
requires both an early and a late reference point, and the two heads
supply these endpoints in a query-dependent division of labor, so neither
head alone carries the difference signal. However, we note that T.h2 exhibits a similar early-key preference, but it enters the circuit only at
$\tau=0.9$ (Table~\ref{tab:circuits}), suggesting possible redundancy among heads anchored to early frames.

\end{document}